\def\year{2020}\relax
\documentclass[letterpaper]{article} % DO NOT CHANGE THIS
\usepackage{aaai20}  % DO NOT CHANGE THIS
\usepackage{times}  % DO NOT CHANGE THIS
\usepackage{helvet} % DO NOT CHANGE THIS
\usepackage{courier}  % DO NOT CHANGE THIS
\usepackage[hyphens]{url}  % DO NOT CHANGE THIS
\usepackage{graphicx} % DO NOT CHANGE THIS
\usepackage{graphicx}  % DO NOT CHANGE THIS
\usepackage{bm}
\usepackage{amssymb}
\usepackage{threeparttable}
\usepackage{epsfig}
\usepackage{graphicx}
\usepackage{amsmath}
\usepackage{amssymb}
\usepackage{multirow}
\usepackage{mathrsfs}
\usepackage{algorithm} %format of the algorithm 
\usepackage{algorithmic} %format of the algorithm 
\usepackage{xcolor}
\title{Decoupled Attention Network for Text Recognition}
\author{
Tianwei Wang,\textsuperscript{\rm 1} 
Yuanzhi Zhu,\textsuperscript{\rm 1} 
Lianwen Jin,\textsuperscript{\rm 1}\thanks{Corresponding author} 
Canjie Luo,\textsuperscript{\rm 1} 
Xiaoxue Chen,\textsuperscript{\rm 1}\\
\Large \textbf{
Yaqiang Wu,\textsuperscript{\rm 2} 
Qianying Wang,\textsuperscript{\rm 2} 
Mingxiang Cai\textsuperscript{\rm 2}}\\
\textsuperscript{\rm 1}School of Electronic and Information Engineering, South China University of Technology \\
\textsuperscript{\rm 2}Lenovo Research \\
wangtw@foxmail.com, z.yuanzhi@foxmail.com, eelwjin@scut.edu.cn, canjie.luo@gmail.com, \\ xxuechen@foxmail.com, wuyqe@lenovo.com, wangqya@lenovo.com, caimx@lenovo.com
}
\begin{document}

\maketitle

\begin{abstract}
Text recognition has attracted considerable research interests because of its various applications.
The cutting-edge text recognition methods are based on attention mechanisms. However, most of attention methods usually suffer from serious alignment problem due to its recurrency alignment operation, where the alignment relies on historical decoding results.
To remedy this issue, we propose a decoupled attention network (DAN), which decouples the alignment operation from using historical decoding results. 
DAN is an effective, flexible and robust end-to-end text recognizer,
which consists of three components: 
1) a feature encoder that extracts visual features from the input image;
2) a convolutional alignment module that performs the alignment operation based on visual features from the encoder; and
3) a decoupled text decoder that makes final prediction by jointly using the feature map and attention maps.
Experimental results show that DAN achieves state-of-the-art performance on multiple text recognition tasks, including offline handwritten text recognition and regular/irregular scene text recognition.
Codes will be released.$\footnote{https://github.com/Wang-Tianwei/Decoupled-attention-network}$
\end{abstract}

\section{Introduction}

Text recognition has drawn much research interest in recent years. 
Benefiting from the development of deep learning and sequence-to-sequence learning, many text recognition methods have achieved notable success \cite{long2018scene}. 
Connectionist temporal classification (CTC) \cite{Graves2006Connectionist} and attention mechanism \cite{bahdanau2015neural} are two most popular methods, among them attention mechanism shows significant better performance and has been studied frequently in recent years \cite{long2018scene}.

\begin{figure}[t]
\centering
\includegraphics[width=0.47\textwidth]{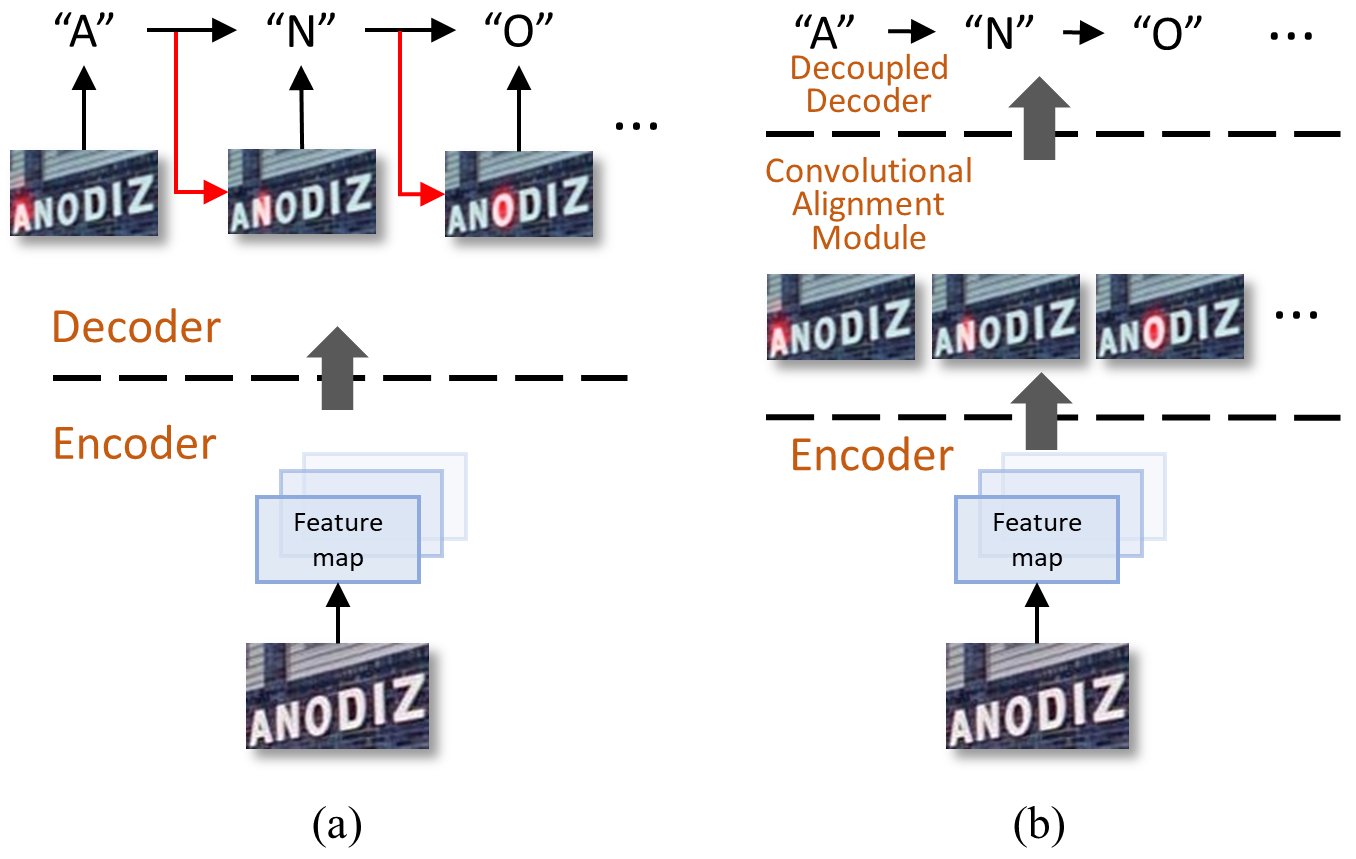}
\caption{(a) Traditional attentional text recognizer, where the alignment operation is conducted using visual information and historical decoding information (red arrow). (b) Decoupled attention network, where the alignment operation is conducted using only visual information.}
\label{Figure_SEQandPAL}
\end{figure}

The attention mechanism, proposed in \cite{bahdanau2015neural} to tackle machine translation problem, was used to handle scene text recognition in \cite{lee2016recursive,shi2016robust}, and since then it dominated text recognition with the following developments \cite{yang2017learning,cheng2017focusing,bai2018edit,cluo2019moran,li2019show}.
The attention mechanism in text recognition is used to align and recognize characters, where the alignment operation has always been coupled with the decoding operation in previous work \cite{shi2016robust,cheng2017focusing,bai2018edit,li2019show}.
As shown in Figure~\ref{Figure_SEQandPAL} (a), the alignment operation of traditional attention mechanism is carried out using two types of information. The first is a feature map that can be regarded as visual information from the encoder, and the second is historical decoding information (in the form of a recurrent hidden state \cite{bahdanau2015neural,luong2015effective} or the embedding vector of previous decoding result \cite{Gehring2017Convolutional,vaswani2017attention}).
The main idea underlying the attention mechanism is matching.
Given a feature from the feature map, its attention score is computed by scoring how well it matches with the historical decoding information \cite{bahdanau2015neural}.

\begin{figure}[t]
\begin{center}
\includegraphics[width=0.47\textwidth]{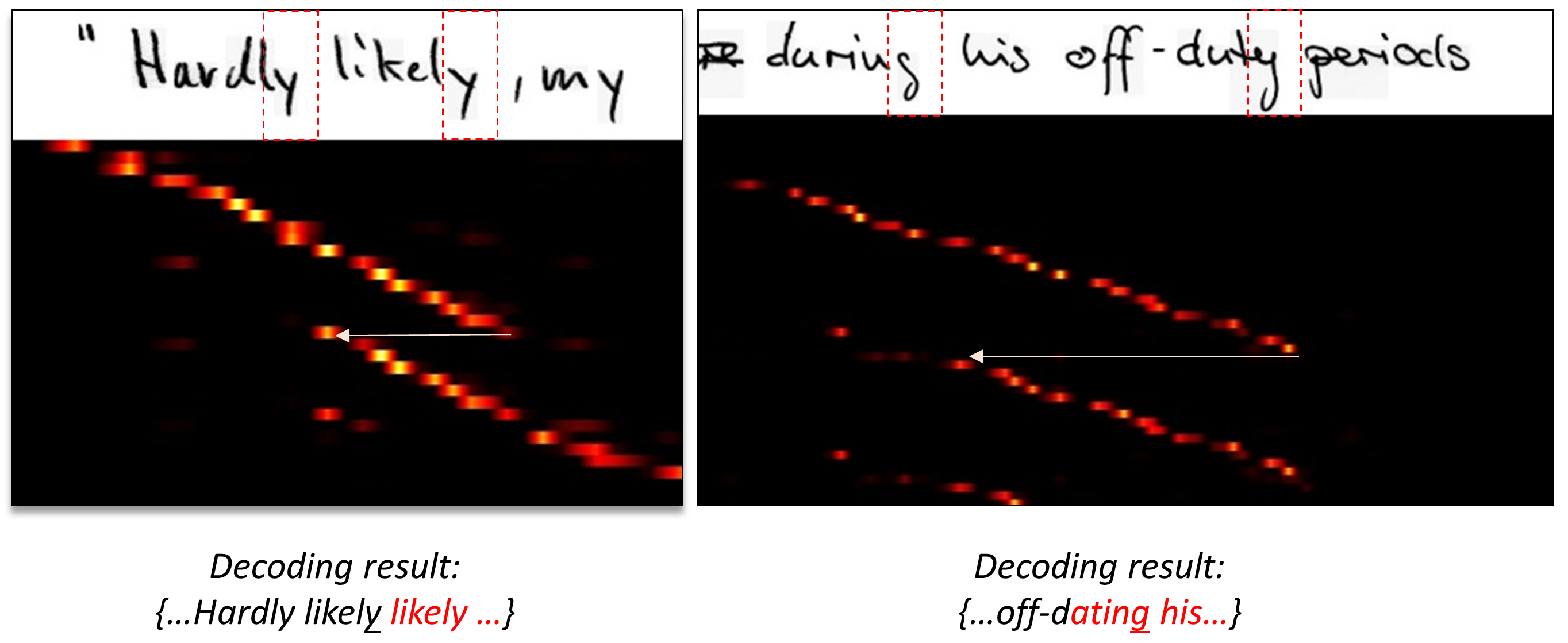}
\caption{Visualization of fractional alignment of traditional attention mechanism \cite{bahdanau2015neural,shi2016robust} on long text.}
\label{Figure_generalattention}
\end{center}
\end{figure}

Traditional attention mechanism often encounters serious alignment problem \cite{cheng2017focusing,bai2018edit,chorowski2015attention-based,kim2017joint}, 
This is because the coupling relationship inevitably leads to error accumulation and propagation.
As shown in Figure~\ref{Figure_generalattention}, the matching-based alignment is easily affected by decoding result.
In the left image, the two consecutive \emph{"ly"} confuses matching operation; in the right image, the misrecognized result \emph{"ing"} confuses matching operation.
\cite{kim2017joint,chorowski2015attention-based} also observed that attention mechanism struggles to align long sequence.
Thus, it is intuitive to find a way to decouple the alignment operation from the historical decoding information, so that to reduce its negative impact.

To solve the aforementioned misalignment issue, in this paper we decouple the decoder of the traditional attention mechanism into an alignment module and a decoupled text decoder, and propose a new method called decoupled attention network (DAN) for text recognition.
As shown in Figure~\ref{Figure_SEQandPAL} (b), compared with traditional attentional scene text recognizer, DAN needs no feedback from the decoding stage for alignment, thus avoiding the accumulation and propagation of decoding errors.
The proposed DAN consists of three components including a feature encoder, a convolutional alignment module (CAM) and a decoupled text decoder.
The feature encoder based on the convolutional neural network (CNN) extracts visual features from the input image.
The CAM, substituting the traditional score-based recurrency alignment module, takes multi-scale visual features from the feature encoder as input, and generates attention maps with a fully convolutional network \cite{long2014Fully} (FCN) in channel-wise manner.
The decoupled text decoder makes the final prediction by using the feature map and attention maps with a gated recurrent unit (GRU) \cite{cho2014properties}.

In summary, our contributions are summarized as follows:
\begin{itemize} %enumerate
\item We propose a CAM to replace the recurrency alignment module in traditional attention decoders. The CAM conducts alignment operation from visual perspective, avoiding the use of historical decoding information, thus eliminating misalignment caused by decoding errors.
\item We propose DAN, which is a effective, flexible (can be easily switched to adapt to different scenarios) and robust (more robust to text length variation and subtle disturbances) attentional text recognizer. 
\item DAN delivers state-of-the-art performance on several text recognition tasks, including handwritten text recognition and regular/irregular scene text recognition.
\end{itemize}

\begin{figure*}[h]
\centering
\includegraphics[width=1.95\columnwidth]{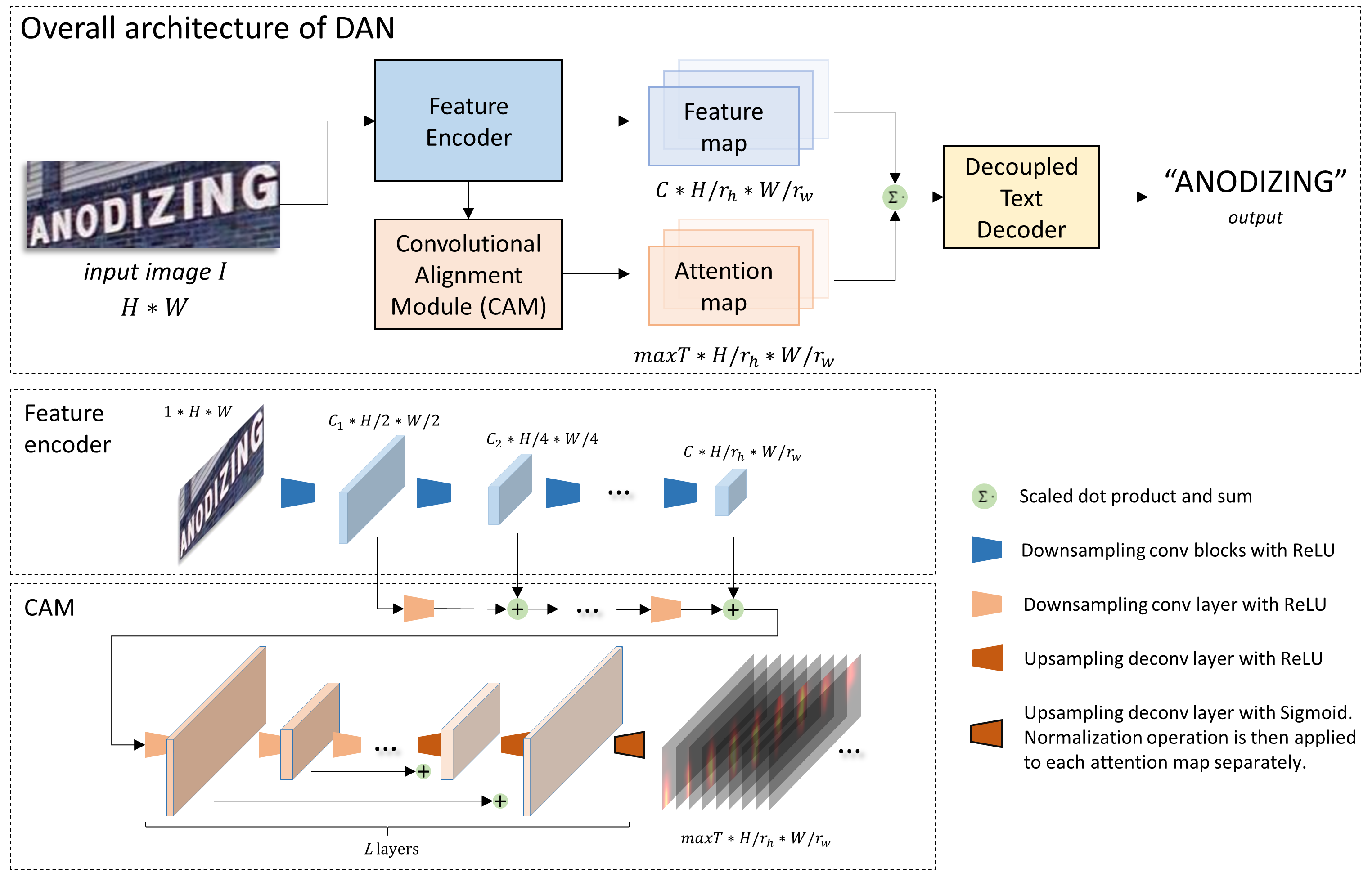}
\caption{Overall architecture of DAN, and detailed architectures of the feature encoder and the CAM. The input image has a normalized height of $H$ and a scaled width of $W$, $C_1$ and $C_2$ are the numbers of channels of the feature map.}
\label{Figure_overall}
\end{figure*}

\section{Related Work}
Text recognition has attracted much research interest in the computer vision community. Early work of scene text recognition relied on low-level features, such as histogram of oriented gradients descriptors \cite{wang2011end}, connected components \cite{neumann2012real}, etc. 
With the rapid development of deep learning, a large number of effective methods have been proposed. These methods can be mainly divided into two branches.

One branch is based on segmentation, it first detects characters then integrates characters into the output.
\cite{bissacco2013photoocr} proposed a five hidden layers for character recognition and a n-gram approach for language modeling.
\cite{wang2012end} used a CNN to recognize characters and adopt a non-maximum suppression to obtain the final predictions.
\cite{jaderberg2014deep} proposed a weight-shared CNN for unconstrained text recognition.
All of these methods require accurate individual detection of characters, which is very challenging.

The other branch is segmentation-free, it recognizes the text line as a whole and focuses on mapping the entire image directly to a word string.
\cite{jaderberg2016reading} regraded scene text recognition as a 90k-class classification task.
\cite{shi2017end} modeled scene text recognition as a sequence problem by integrating the advantages of both deep convolutional neural network and recurrent neural network, and CTC was used to train the model end-to-end.
\cite{lee2016recursive} and \cite{shi2016robust} introduced attention mechanism to automatically align and translate words.
From then on, more and more attention-based methods were proposed for text recognition.
\cite{cheng2017focusing} observed the attention drift problem and proposed a focusing net to draw back the drifted attention, but character-level annotation was required.
\cite{bai2018edit} proposed a post-process, the edit probability to re-estimate the alignment; but they did not fundamentally solve misalignment.
Focusing on recognition of irregular text, \cite{shi2016robust}, \cite{cluo2019moran} and \cite{Zhan2019ESIR} proposed to rectify text distortion and recognize the rectified text with an attention-based recognizer; \cite{liu2018char} proposed to rectify text at the character level; \cite{yang2017learning} and \cite{liao2019scene} proposed to recognize text in two-dimensional perspective but character-level annotation is required; \cite{cheng2018aon} proposed to capture character feature in four directions.
\cite{fang2018attention} proposed an attention and language ensemble network, and multiple losses from attention and language are accumulated for training it.
\cite{li2019show} proposed a simple and effective model using 2D attention mechanism.

Despite the notable success achieved by these attention-based methods, all of them consider attention to be a coupled operation between historical decoding information and visual information, and no study to date has focused on applying attention mechanism in long text recognition to the best of our knowledge.

\section{DAN}
The proposed DAN aims at solving the misalignment issue of traditional attention mechanism through decoupling the alignment operation from using historical decoding results. To this end, we proposed a new convolutional alignment module (CAM) together with a decoupled text decoder to replace the traditional decoder. The overall architecture of DAN is illustrated in Figure~\ref{Figure_overall}. Details will be introduced in the followings.

\subsection{Feature Encoder}
We adopt a similar CNN-based feature encoder as previous study \cite{shi2018aster}.
The feature encoder $\mathcal{F}$ encodes the input image $\bm{x}$ of size $H \times W$ into feature map $\bm{F}$:
\begin{align}
\bm{F}=\mathcal{F}(\bm{x}), \bm{F}\in\mathcal{R}^{C \times H/r_h \times W/r_w}. 
\end{align}
where $C$ , $r_h$ and $r_w$ denote the output channels, the height and the width downsampling ratio respectively.

\subsection{Convolutional Alignment Module (CAM)}
As shown in Figure~\ref{Figure_overall}, the input of our proposed CAM is visual features of each scale from the feature encoder. These multi-scale features are first encoded by cascade down-sampling convolutional layers then summarized as input.
Inspired by the FCN that makes dense predictions per-pixel channel-wise (\emph{i.e}., each channel denotes a heatmap of a class), we use a simple FCN architecture to conduct the attention operation channel-wise, which is quite different from current attention mechanism.
The CAM has $L$ layers; in the deconvolution stage, each output feature is added with the corresponding feature map from convolution stage. Sigmoid function with channel-wise normalization is finally adopted to generate attention maps $\bm{A} = \{\bm{\alpha_1}, \bm{\alpha_2}, ..., \bm{\alpha_{maxT}}\}$, where $maxT$ denotes the maximum number of channels, \emph{i.e}., the maximum number of decoding steps; and the size of each attention map is $H/r_h \times W/r_w$.

Compared with the FCN used for semantic segmentation, the CAM plays a completely different role to model a sequential problem. Although $maxT$ is pre-defined and should be fixed during training and testing, we will experimentally show that the setting of $maxT$ does not influence the final performance as long as it is reasonable. 

\begin{figure}[h]
\centering
\includegraphics[width=0.4\textwidth]{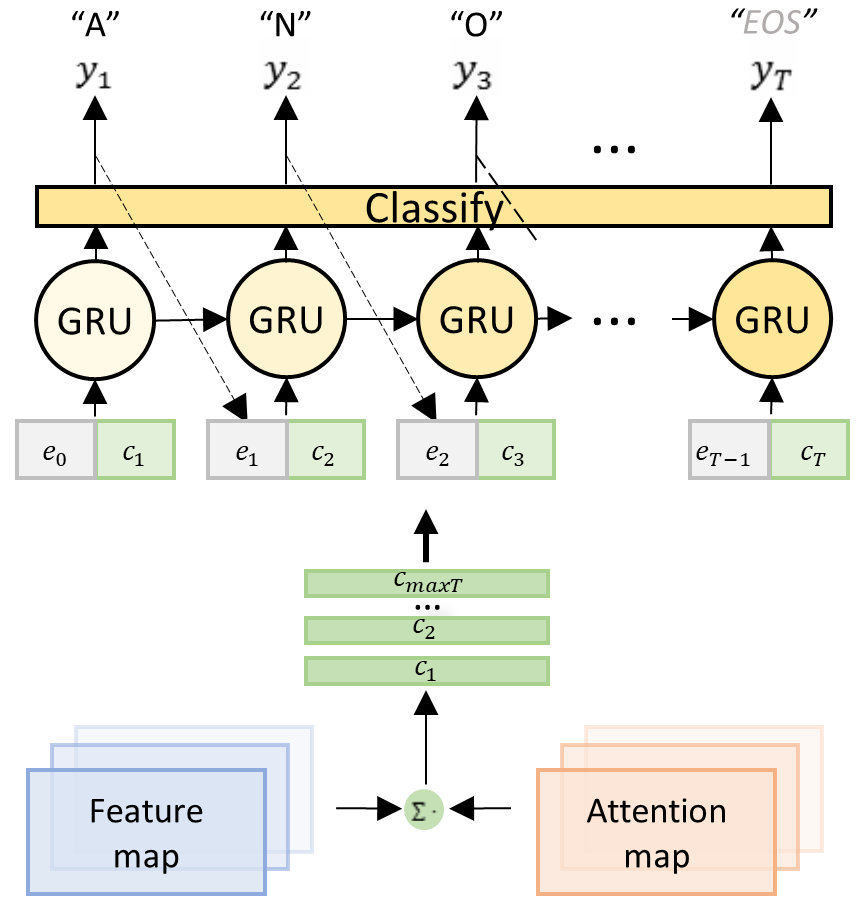}
\caption{Detailed architecture of the decoupled text decoder. It consists of a GRU layer used to explore the contextual information and a linear layer to make predictions. `\emph{EOS}' denotes end-of-sequence symbol.}
\label{Figure_textdecoder}
\end{figure}

By controlling the downsampling ratio $r_h$ and change the stride of CAM, DAN can be flexibly switched between 1D and 2D form.
When $H/r_h=1$, DAN becomes a 1D recognizer and is suitable for long and regular text recognition; When $H/r_h>1$(\emph{e.g}., for input image with height of 32, $r_h=4$ results in a feature map with height of 4), DAN becomes a 2D recognizer and is suitable for irregular text recognition.
Compared with previous 2D scene text recognizers, \cite{yang2017learning,liao2019scene} which need character-level annotation for supervision; \cite{li2019show} which uses a tailored 2D attention for 2D spatial relationships caption, result in more complex than 1D form and has a poor performance on regular text recognition, DAN is significantly simple and flexible, while achieves state-of-the-art or comparable performance both in 1D (handwritten text) and 2D (irregular scene text) recognition.

\subsection{Decoupled Text Decoder}
Different from the traditional attentional decoder that conduct alignment and recognition concurrently, our decoupled text decoder takes encoded features and attention maps as input, and conducts recognition only.
As shown in Figure~\ref{Figure_textdecoder}, the decoupled text decoder computes context vector $c_t$ as:
\begin{align}
c_t = \sum^{W/r_w}_{x=1}\sum^{H/r_h}_{y=1}\alpha_{t,x,y}F_{x,y}.
\end{align}
At time step $t$, the classifier generates output $y_t$:
\begin{align}
y_t = wh_t + b,
\end{align}
where $h_t$ is the hidden state of the GRU, computed as:
\begin{align}
h_t = GRU((e_{t-1}, c_t), h_{t-1}),
\end{align}
$e_t$ is an embedding vector of the previous decoding result $y_t$.
The loss function of DAN is as follows:
\begin{align}
Loss = -\sum^{T}_{t=1}logP(g_t|I,\theta),
\end{align}
where $\theta$ and $g_t$ denote all trainable parameters in the DAN and groudtruth at step $t$, respectively.
Just like other attentional text recognizers, DAN uses word-level annotation for training.

\begin{table}[htb]
\caption{Detailed configuration of the feature encoder. `Num' and `hw' mean number of blocks and handwritten text recognition experiments, respectively.}
\centering
\label{Table_featureencoder}
\resizebox{.98\columnwidth}{!}{
\begin{tabular}{c|c|c|ccc}
\hline
\multirow{2}{*}{\textbf{Name}} & \multirow{2}{*}{\textbf{Configuration}} & \multirow{2}{*}{\textbf{Num}} & \multicolumn{3}{c}{\textbf{Downsampling Ratio}} \\ \cline{4-6}
& & & hw & scene-1D & scene-2D \\ \hline
Res-block0 & 3 $\times$ 3 conv & 1 & 2$\times$1 & 1$\times$1 & 1$\times$1 \\ \hline
Res-block1 & \begin{tabular}[c]{@{}c@{}}1 $\times$ 1 conv, 32\\ 3 $\times$ 3 conv, 32\end{tabular}  & 3 & 2$\times$2 & 2$\times$2 & 2$\times$2 \\ \hline
Res-block2 & \begin{tabular}[c]{@{}c@{}}1 $\times$ 1 conv, 64\\ 3 $\times$ 3 conv, 64\end{tabular}  & 4 & 2$\times$2 & 2$\times$2 & 1$\times$1 \\ \hline
Res-block3 & \begin{tabular}[c]{@{}c@{}}1 $\times$ 1 conv, 128\\ 3 $\times$ 3 conv, 128\end{tabular} & 6 & 2$\times$1 & 2$\times$1 & 2$\times$2 \\ \hline
Res-block4 & \begin{tabular}[c]{@{}c@{}}1 $\times$ 1 conv, 256\\ 3 $\times$ 3 conv, 256\end{tabular}  & 6 & 2$\times$2 & 2$\times$1 & 1$\times$1 \\ \hline
Res-block5 & \begin{tabular}[c]{@{}c@{}}1 $\times$ 1 conv, 512\\ 3 $\times$ 3 conv, 512\end{tabular}   & 3 & 2$\times$2 & 2$\times$1 & 1$\times$1 \\ \hline
\end{tabular}
}
\end{table}

\section{Performance Evaluation}
In our experiments, two tasks are employed to evaluate the effectiveness of DAN, including handwritten text recognition and scene text recognition.
The detailed network configuration of feature encoder is given in Table~\ref{Table_featureencoder}.

\subsection{Offline Handwritten Text Recognition}
Owing to its long sentences (up to 90 characters), diverse writing styles, and character-touching problem, the offline handwritten text recognition problem is highly complicated and challenging to solve. 
Therefore, it is a favorable testbed to evaluate the robustness and effectiveness of DAN.

For exhaustive comparison, we also conduct experiments on two popular attentional decoders: Bahdanau's attention \cite{bahdanau2015neural} and Luong's attention \cite{luong2015effective}. These attentional decoders are widely adopted for text recognition \cite{shi2018aster,cheng2018aon,cluo2019moran,li2019show}.
When comparing with these decoders, the CAM and decoupled text decoder are replaced by them for the sake of fairness.

\subsubsection{Datasets}
Two public handwritten datasets are used to evaluate the effectiveness of DAN, including IAM \cite{Marti2002The} and RIMES \cite{Grosicki2009Results}.
The IAM dataset is based on handwritten English text copied from the LOB corpus. It contains 747 documents (6,482 lines) in the training set, 116 documents (976 lines) in the validation set and 336 documents (2,915 lines) in the test set. 
The RIMES dataset consists of handwritten letters in French. There are 1,500 paragraphs (11,333 lines) in the training set, and 100 paragraphs (778 lines) in the testing set. 

\subsubsection{Implementation Details}
On both databases we use the original whole-line training set with an open-source data-augmentation toolkit$\footnote{https://github.com/Canjie-Luo/Scene-Text-Image-Transformer}$ to train the network.
The height of the input image is normalized as 192 and the width is calculated with the original aspect ratio (up to 2048).
To downsample the feature map into 1D, we add a convolution layer with kernel size 3$\times$1 to the end of the feature encoder.
$maxT$ is set to 150 in order to cover the longest line.
The measure of performance is the Character or Word Error Rate (CER\% or WER\%), corresponding to the edit distance between the recognition result and groundtruth, normalized by the number of groundtruth characters (or words).
At test time on RIMES dataset, we crop the test image with six pre-defined strategies (\emph{e.g}., \{10,10\} meant that the top 10 rows and the bottom 10 rows are cropped out), and then conduct recognition on them and the original image. A recognition score is calculated by averaging the output probabilities and the top scored one is chosen as the final result.
All the layers of CAM except the last one are set as 128 channels in order to cover the longest text length.
No language model or lexicon is used during experiments.

\begin{table}[h]
\caption{Performance comparison on handwritten text datasets.}
\label{Table_handwritten}
\begin{center}
\begin{threeparttable}
\begin{tabular}{|c|c|c|c|c|}
\hline
\multirow{2}{*}{\textbf{Methods}} & \multicolumn{2}{c|}{\textbf{IAM}} & \multicolumn{2}{c|}{\textbf{RIMES}} \\ \cline{2-5}
& \textbf{WER} & \textbf{CER} & \textbf{WER} & \textbf{CER} \\ \hline
\cite{Salvador2011Improving} & 22.4  & 9.8 & - & - \\ 
(Pham et al. 2014) & 35.1 & 10.8 & 28.5 & 6.8 \\
\cite{Bluche2016Joint} & 24.6 & 7.9 & 12.6 & 2.9 \\ 
\cite{sueiras2018offline} & 23.8 & 8.8 & 15.9 & 4.8 \\ 
\cite{bhunia2019handwriting}\tnote{1} & \textbf{17.2} & 8.4 & 10.5 & 6.4 \\
\cite{zhang2019sequence-to-sequence} & 22.2 & 8.5 & - & - \\ \hline\hline
DAN & 19.6 & \textbf{6.4} & \textbf{8.9} & \textbf{2.7} \\ \hline
\end{tabular}
\begin{tablenotes}
\item[1] Word-level recognition, where the words in the original image are cropped out then recognized.
\end{tablenotes}
\end{threeparttable}
\end{center}
\end{table}
\subsubsection{Experimental Results}
As shown in Table~\ref{Table_handwritten}, DAN exhibits superior performance on both datasets.
On IAM dataset, DAN outperforms previous state-of-the-art by 1.5\% on CER.
Note that although \cite{bhunia2019handwriting} shows better performance on WER, their method needs cropped word images as input, while our method directly recognizes text lines.
On RIMES, it is inferior to previous state-of-the-art by 0.2\% on CER; but on WER, it has a great error reduction of 3.7\% (relative error reduction of 29\%).
The great improvement in terms of WER indicates that DAN has a stronger capability of learning semantic information, which is helpful for long text recognition.

\begin{table}[t]
\caption{Performance comparison on different output lengths. The `time/iter' means forward time per iteration on TITAN X GPU.}
\label{Table_outputcomparison}
\begin{center}
\begin{tabular}{|c|c|c|c|}
\hline
\multirow{2}{*}{\textbf{output length}} & \multicolumn{2}{c|}{\textbf{IAM}} & \multirow{2}{*}{\textbf{time/iter}}\\ \cline{2-3}
& WER & CER & \\ \hline
150 & 19.6 & 6.4 & 188.7 ms \\ 
200 & 19.5 & 6.3 & 189.5 ms \\ 
250 & 19.6 & 6.4 & 190.5 ms \\ \hline
\end{tabular}
\end{center}
\end{table}

\begin{figure}[h]
\centering
\includegraphics[width=0.4\textwidth]{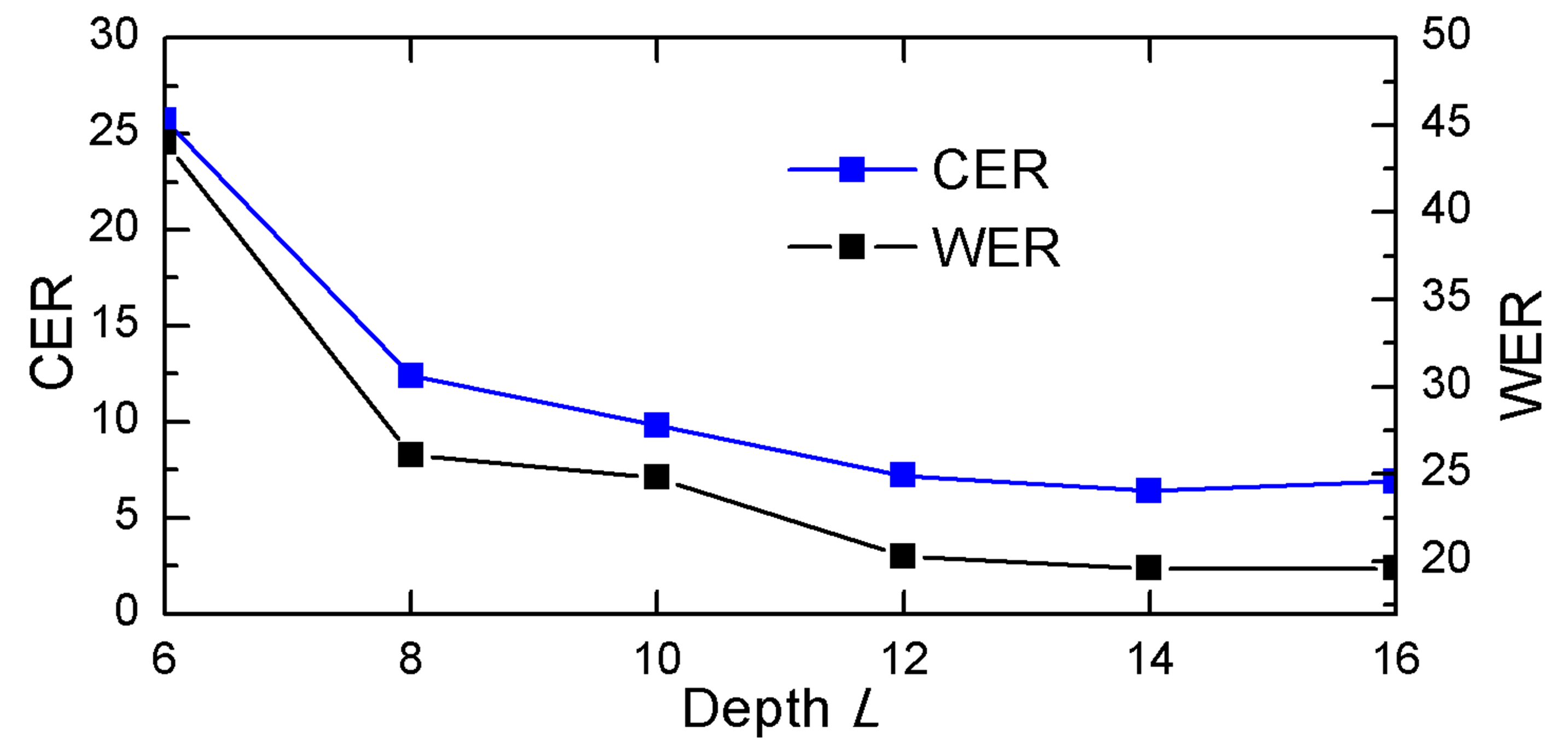}
\caption{Performance comparison of different depth $L$ on IAM dataset.}
\label{Figure_depthL}
\end{figure}

\begin{table}[h]
\caption{Performance comparison of different decoders. FE denotes the feature encoder of DAN. `Bah' and `Luong' denote Bahdanau's attention and Luong's attention, respectively.}
\label{Table_decoders}
\begin{center}
\begin{tabular}{|c|c|c|c|c|}
\hline
\multirow{2}{*}{\textbf{Methods}} & \multicolumn{2}{c|}{\textbf{IAM}} & \multicolumn{2}{c|}{\textbf{RIMES}} \\ \cline{2-5}
& \textbf{WER} & \textbf{CER} & \textbf{WER} & \textbf{CER} \\ \hline
FE + Bah & 25.9 & 9.9 & 9.1 & 3.0 \\ 
FE + Luong & 25.7 & 10.3 & 9.3 & 3.3 \\ 
DAN & \textbf{19.6} & \textbf{6.4} & \textbf{8.9} & \textbf{2.7} \\ \hline
\end{tabular}
\end{center}
\end{table}

\begin{figure}[h]
\centering
\includegraphics[width=0.45\textwidth]{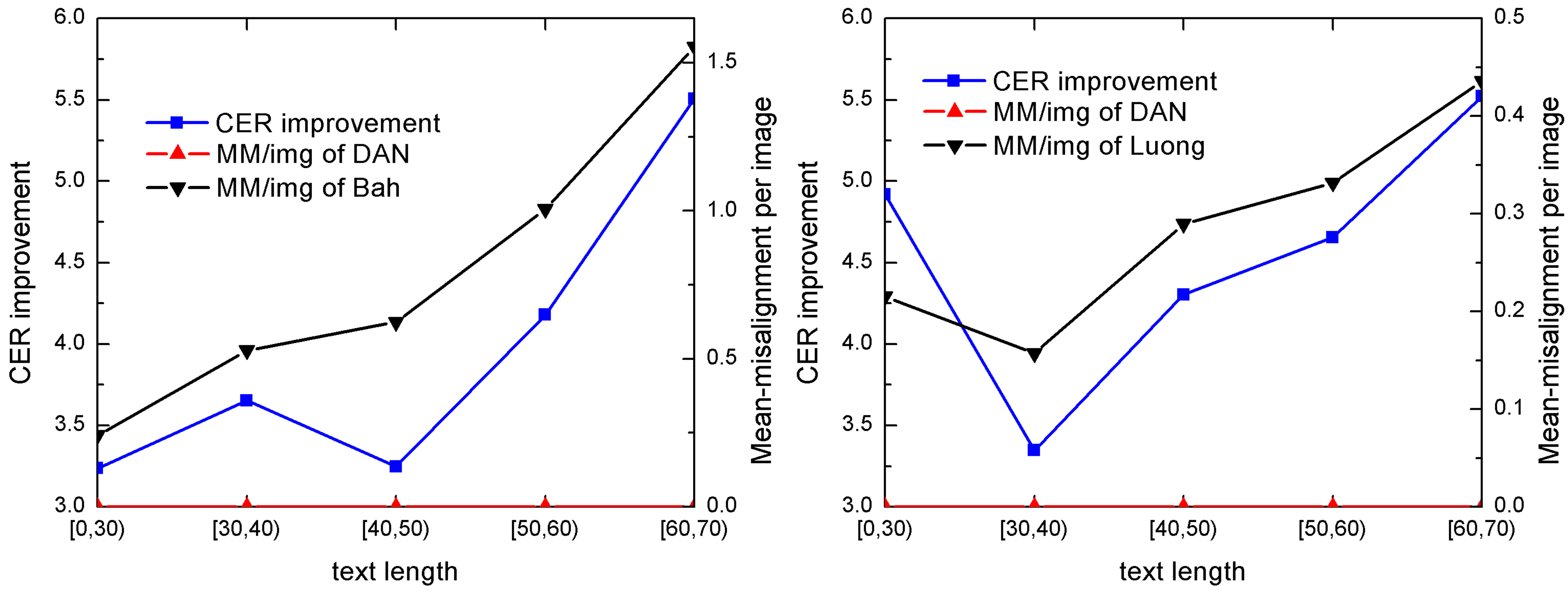}
\caption{CER improvements of DAN on different text lengths and corresponding misalignments. `Bah' and `Luong' denote Bahdanau's attention and Luong's attention, respectively.}
\label{Figure_alignCERimprov}
\end{figure}

\subsubsection{Ablation Study}
In this subsection, we will evaluate the influence of different depth $L$ and output length $maxT$ of CAM.

\textbf{Output length}: As shown in Table~\ref{Table_outputcomparison}, different output lengths do not influence the performance, and the computation resource of additional channels is negligible, which indicates that DAN works well as long as the output length is reasonably set (longer than text length). 

\textbf{Depth}: As shown in Figure~\ref{Figure_depthL}, the performance of DAN degrades seriously as we reduce $L$, which show that the CAM should be deep enough to reach good performance.
To successfully align one character, the reception field of CAM must be big enough to cover the corresponding features of this character and its neighbor regions.

\subsubsection{Deep Insight into Eliminating Misalignments}
As shown in Table~\ref{Table_decoders}, compared with these two widely-used attentional decoders in the field of text recognition, DAN achieves significantly better performance.

To fine-grained study the improvements brought by the better alignment of DAN, we quantitatively discuss the relationship between obtained improvements of DAN and corresponding eliminated alignment errors.
We propose a simple misalignment measurement method, which is based on the priori knowledge that all texts are written from left to right.
This method consists of two steps: 1) picking the region with maximum attention score as attention center; 2) if current attention center is on the left side of the previous one, recording one misalignment.
We divide the test samples into five groups by the text length: $[0,30), [30,40), [40,50), [50,60), [60,70)$; each group contains more than 100 samples.
In each group, the misalignments are added up then averaged to produce mean-misalignments per image (MM/img).

\begin{figure}[h]
\centering
\includegraphics[width=0.45\textwidth]{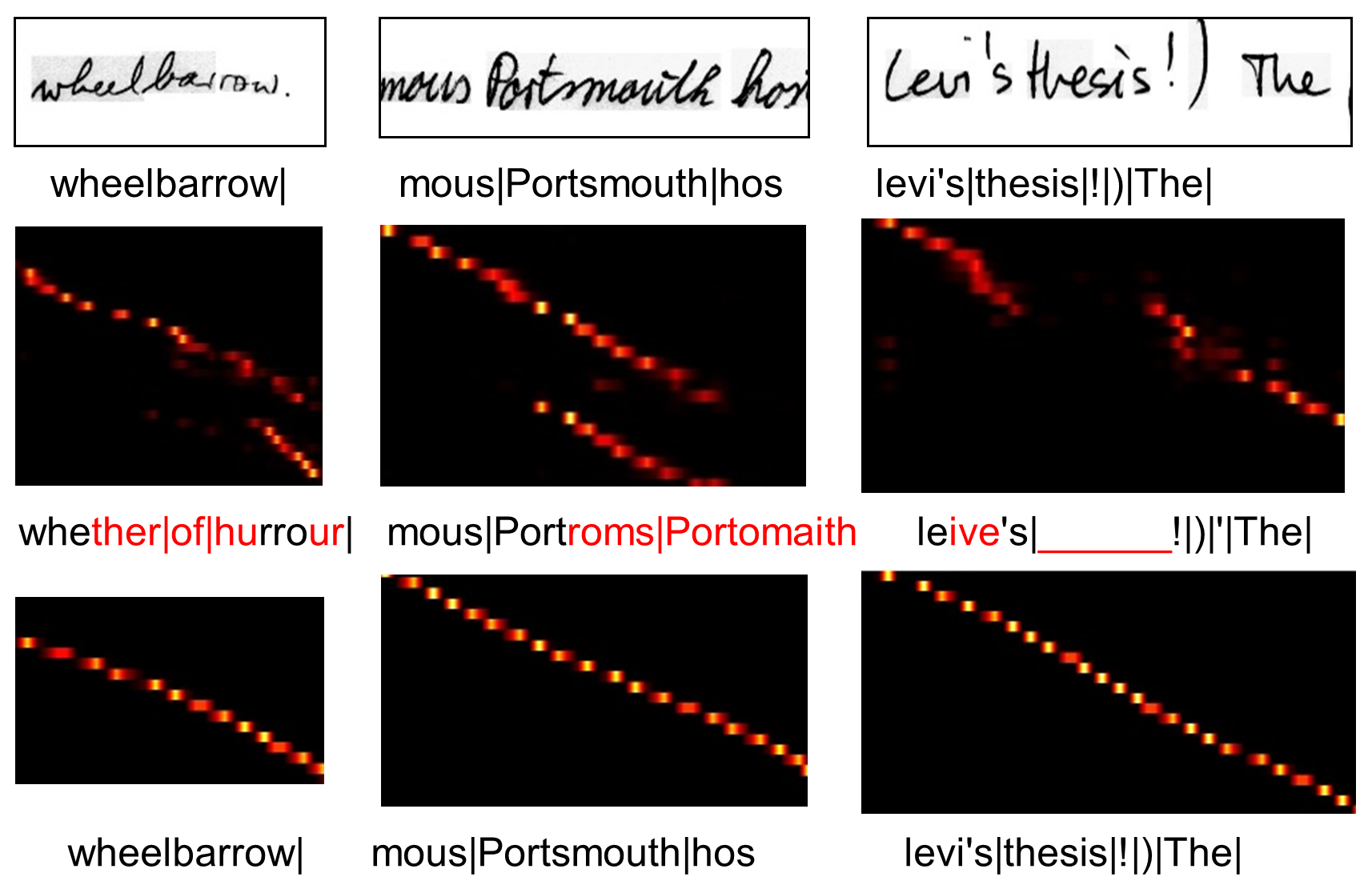}
\caption{Visualization of attention maps and recognition results on IAM dataset. Top: original fractional images and corresponding groundtruth; middle: attention maps and recognition results of traditional attention; bottom: attention maps and recognition results of DAN.}
\label{Figure_IAM_compare}
\end{figure}

The experimental results are shown in Figure~\ref{Figure_alignCERimprov}; The changes of CER improvement and eliminated misalignments are almost the same trend, which validates the performance gain of DAN relative to traditional attention comes from eliminating misalignments. 
In Figure~\ref{Figure_IAM_compare}, we show some visualization results of eliminated misalignments by our DAN. 

\subsubsection{Error Analysis}
\begin{figure}[h]
\centering
\includegraphics[width=0.45\textwidth]{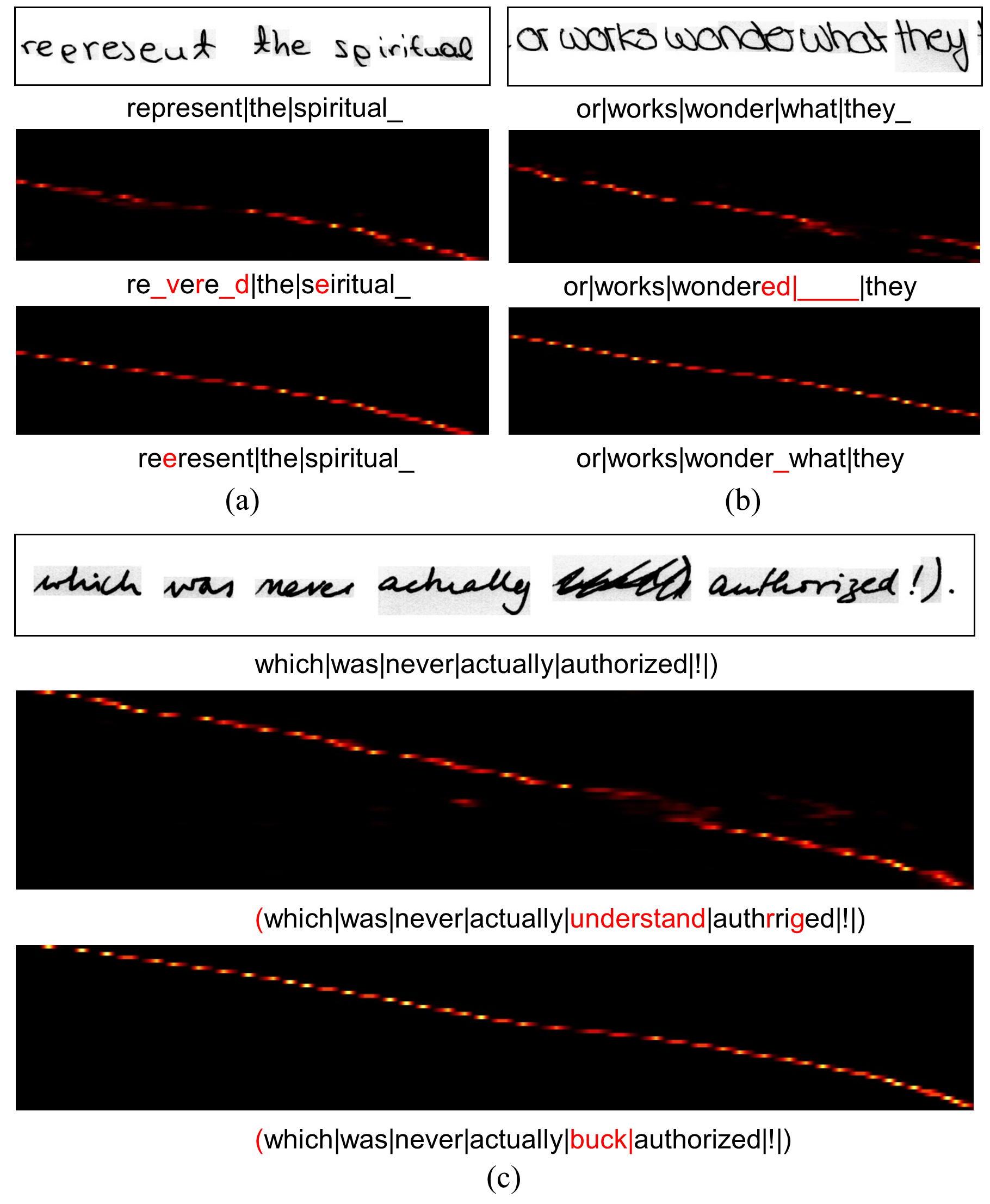}
\caption{Visualization of typical error samples of DAN. The order of images is same as Figure~\ref{Figure_IAM_compare}. (a) Substitute error where character `p' is misrecognized as `e'; (b) delete error where a space symbol is missed; (c) insert error where some textures are recognized as `buck'.}
\label{Figure_error_sample}
\end{figure}

Figure~\ref{Figure_error_sample} shows some typical error samples of DAN.
In Figure~\ref{Figure_error_sample} (a), the character `e' is recognized as `p' because of its confusing writing style. 
The misclassified `p' is challenging for humans without contextual information.
In Figure~\ref{Figure_error_sample} (b), a space symbol is missed by the recognizer, because the two relevant words are too close. 
In Figure~\ref{Figure_error_sample} (c), some noise texture is recognized as a word by DAN.
However, DAN is still more robust than traditional attention on these samples. 
In Figure~\ref{Figure_error_sample} (c) the confusing noises disturb the alignment operation of traditional attention and lead to unpredictable errors, while DAN is robust in alignment even if extra results are generated. 
Considering that the noises have almost the same texture with normal text, this type of error is very difficult to avoid, especially for DAN which conduct alignment only based on visual features.

\begin{table*}[h]
\caption{Performance comparison on regular and irregular scene text datasets. `Rect' represents rectification-based methods; `2D' represents 2D-based methods.}
\label{Table_scene}
\begin{center}
\begin{threeparttable}
\begin{tabular}{|c|c|c|cccc|ccc|}
\hline
\multirow{2}{*}{\textbf{Methods}} & \multirow{2}{*}{\textbf{\textbf{Rect}}} & \multirow{2}{*}{\textbf{\textbf{2D}}} & \multicolumn{4}{c|}{\textbf{Regular}} & \multicolumn{3}{c|}{\textbf{Irregular}} \\\cline{4-10}
& & &\textbf{IIIT5k} & \textbf{SVT} & \textbf{IC03} & \textbf{IC13} & \textbf{SVT-P} & \textbf{CUTE80} & \textbf{IC15} \\ \hline
\cite{cheng2017focusing} \tnote{1} &  &  & 87.4  & 85.9  & 94.2   & 93.3 & - & - & 70.6 \\
\cite{cheng2018aon} &  &  & 87.0  & 82.8  & 91.5   & -   & 73.0  & 76.8 & 68.2  \\
\cite{bai2018edit}  \tnote{1} & &  & 88.3  & 87.5  & 94.6 & \textbf{94.4} & - & - & 73.9\\
\cite{liu2018synthetically} & & & 89.4  & 87.1  & 94.7   & 94.0 & 73.9  & 62.5 & -  \\
\cite{shi2018aster} & \checkmark & & 93.4  & 89.5 & 94.5 & 91.8 & 78.5   & 79.5   & 76.1  \\ 
\cite{fang2018attention} & & & 86.7 & 86.7 & 94.8 & 93.5 & -  & - & 71.2 \\
\cite{cluo2019moran}& \checkmark &  & 91.2 & 88.3  & 95.0 & 92.4 & 76.1 & 77.4   & 68.8 \\
\cite{liao2019scene}  \tnote{1} & & \checkmark & 92.0  & 86.4  & -   & 91.5 \tnote{1} & - & 79.9   & - \\ 
\cite{li2019show} & & \checkmark & 91.5  & 84.5  & -   & 91.0 & 76.4 & 83.3 & 69.2 \\ 
\cite{Xie2019Aggregation} & & \checkmark & - & - & - & - & 70.1 & 82.6 & 68.9 \\ 
\cite{Zhan2019ESIR} & \checkmark & & 93.3 & \textbf{90.2} & - & 91.3 & 79.6 & 83.3 & \textbf{76.9} \\ \hline\hline
DAN-1D  & & & 93.3  & 88.4  & \textbf{95.2} & 94.2 & 76.8  & 80.6 & 71.8  \\
DAN-2D  & & \checkmark & \textbf{94.3}  & 89.2  & 95.0 & 93.9 & \textbf{80.0}  & \textbf{84.4}  & 74.5 \\ \hline
\end{tabular}
\begin{tablenotes}
\item[1] character-level annotation required.
\end{tablenotes}
\end{threeparttable}
\end{center}
\end{table*}

\begin{table*}[h]
\begin{center}
\caption{Robustness study. `ac': accuracy; `gap': the gap between the original dataset; `ratio': accuracy decreasing ratio.}
\label{Table_robustnessstudy}
\begin{tabular}{|c|c|c|c|c|c|c|c|c|c|c|c|c|c|c|}
\hline
\multirow{2}{*}{\textbf{Methods}} & \textbf{IIIT} & \multicolumn{3}{c|}{\textbf{IIIT-p}} & \multicolumn{3}{c|}{\textbf{IIIT-r-p}} & \textbf{IC13} & \multicolumn{3}{c|}{\textbf{IC13-ex}} & \multicolumn{3}{c|}{\textbf{IC13-r-ex}} \\\cline{2-15}
& ac & ac & gap & ratio & ac & gap & ratio & ac & ac & gap & ratio & ac & gap & ratio  \\ \hline
CA-FCN & 92.0 & 89.3 & -2.7 & 2.9\% & 87.6 & \textbf{-4.4} & \textbf{4.8\%} & 91.4 & 87.2 & -3.7 & 4.1\% & 83.8 & \textbf{-6.9} & 7.6\% \\ 
DAN-1D & 93.3 & 91.5 & \textbf{-1.8} & \textbf{1.9\%} & 88.2 & -5.1 & 5.4\% & \textbf{94.2} & \textbf{91.2} & \textbf{-3.0} & \textbf{3.2\%} & \textbf{86.9} & -7.3 & 7.7\% \\
DAN-2D & \textbf{94.3}  & \textbf{92.1} & -2.2 & 2.3\% & \textbf{89.1} & -5.2 & 5.5\% & 93.9 & 90.4 & -3.5 & 3.7\% & \textbf{86.9} & -7.0 & \textbf{7.5\%} \\ \hline
\end{tabular}
\end{center}
\end{table*}

\subsection{Scene Text Recognition}

Scene text recognition often encounters problems owing to the large variations in the background, appearance, resolution, text font, and so on. 
In this section, we will study the effectiveness and robustness of DAN on seven datasets including regular scene text datasets and irregular scene text datasets.
We will validate the performance of DAN in 1D and 2D form (denote as DAN-1D and DAN-2D); the detailed configurations of feature encoder are shown in Table~\ref{Table_featureencoder}.

\subsubsection{Datasets}
Two types of datasets are used for scene text recognition: regular scene text datasets, including IIIT5K-Words \cite{mishra2012scene}, Street View Text \cite{wang2011end}, ICDAR 2003 \cite{lucas2005icdar} and ICDAR 2013 \cite{karatzas2013icdar}; and irregular scene text datasets, including SVT-Perspective \cite{neumann2012real}, CUTE80 \cite{risnumawan2014robust} and ICDAR 2015 \cite{karatzas2015icdar}.

IIIT5k was collected from the Internet, and contained 3,000 cropped word images for testing. 

Street View Text (SVT) was collected from the Google Street View, and contained 647 word images for testing. 

ICDAR 2003 (IC03) contained 251 scene images that are
labeled with text bounding boxes. The dataset contained 867 cropped images.

ICDAR 2013 (IC13) inherited most images from IC03 and extends it with some new images. It consisted of 1,015 cropped images without associated lexicon.

SVT-Perspective (SVT-P) was collected from the side-view angle snapshots in Google Street View, and contained 639 cropped images for testing . 

CUTE80 focused on curved text, and consisted of 80 high-resolution images taken in natural scenes. 
This dataset contained 288 cropped natural images for testing.

ICDAR 2015 (IC15) contained 2,077 cropped images. 
A large proportion of images were blurred and multi-oriented.

\subsubsection{Implementation Details}
We train our model on synthetic samples released by \cite{jaderberg2014synthetic} and \cite{gupta2016synthetic}.
For better comparison, we compare DAN only with the methods that had also used these two synthetic datasets.
The height of the input image is set to 32 and the width is calculated with the original aspect ratio (up to 128).
$maxT$ is set as 25; $L$ is set as 8; and all the layers of CAM except the last one are set as 64.
We use the bi-directional decoder proposed in \cite{shi2018aster} for final prediction.
channels. 
With ADADELTA \cite{zeiler2012adadelta} optimization method, the learning rate is set as 1.0 and reduced to 0.1 after the third epoch.

\subsubsection{Experimental Results}
\label{Sec_scenetextresult}
As shown in Table~\ref{Table_scene}, DAN achieves state-of-the-art or comparable performance on most datasets.
For regular scene text recognition, DAN achieves state-of-the-art performance on IIIT5K and IC03, and is just a little behind the current state-of-the-art on SVT and IC13. 
DAN-1D performs a little better on IC03 and IC13, because images from these two datasets are usually clean and regular.
For irregular scene text recognition, the most advanced methods can be divided into two types: rectification based and 2D based.
DAN-2D achieves state-of-the-art performance on SVT-P and CUTE80, and it exhibits the best performance among 2D recognizers.

\subsubsection{Robustness Study}
Scene text is usually affected by environmental disturbances.
To check whether DAN is sensitive to subtle disturbances, we also conduct robustness study on IIIT-5k and IC13 datasets, and compare DAN with the most-recent 2D scene text recognizer, CA-FCN \cite{liao2019scene}. 
We add some disturbances on these two datasets as follows:

\textbf{IIIT-p}: Padding the images in IIIT5k with extra 10\% height vertically and 10\% width horizontally by repeating the border pixels.
\textbf{IIIT-r-p}: 1. Separately stretching the four vertexes of the images in IIIT5k with a random scale up to 20\% of height and
width respectively. 2. Repeating border pixels to fill the
quadrilateral images. 3. Transforming the images back to axis-aligned rectangles.
\textbf{IC13-ex}: Expanding the bounding boxes of the images in IC13 to expanded rectangles with extra 10\% height and width before cropping.
\textbf{IC13-r-ex}: 1. Expanding the bounding boxes of the images in IC13 randomly with a maximum 20\% of width and height to form expanded quadrilaterals. 2. The pixels in axis-aligned circumscribed rectangles of those images are cropped.

The results are shown in Table~\ref{Table_robustnessstudy}. In most cases DAN exhibits to be more robust than CA-FCN, which again validates its robustness.

\subsection{Discussion}
\subsubsection{Advances of DAN:}
1) \textbf{Simple}. 
DAN uses off-the-shelf components; all of them are easy to implement. 
2) \textbf{Effective}.
DAN achieves state-of-the-art performance on multiple text recognition tasks.
3) \textbf{Flexible}.
The form of DAN can be easily switched between 1D and 2D.
4) \textbf{Robust}.
DAN exhibits more reliable alignment performance when facing long text. It is also more robust facing subtle disturbances.

\subsubsection{Limitations of DAN:}
The CAM uses only visual information for alignment operation; thus when it comes text-like noises, it struggles to align the text.
This kind of error is shown in Figure~\ref{Figure_error_sample} (c) and may be a common issue for most attention mechanism.

\section{Conclusion}
In this paper, an effective, flexible and robust decoupled attention network is proposed for text recognition.
To address the misalignment issue, DAN decouples the decoder of the traditional attention mechanism into a convolutional alignment module and a decoupled text decoder.
Compared with the traditional attention mechanism, DAN effectively eliminates the alignment errors and achieves the state-of-the-art performance.
Experimental results on multiple text recognition tasks have shown its effectiveness and merit.
Particularly, DAN shows significant superiority when dealing with long text recognition, such as handwritten text recognition.  

\section{Acknowledgement}
This research is supported in part by NSFC (Grant No.: 61936003), the National Key Research and Development Program of China (No. 2016YFB1001405), GD-NSF (no.2017A030312006), Guangdong Intellectual Property Office Project (2018-10-1), and GZSTP (no. 201704020134).

{
\bibliographystyle{aaai}
\bibliography{AAAI-WangT.3550}
}

\end{document}